\documentclass[10pt,twocolumn,letterpaper]{article}

\usepackage{wacv}
\usepackage{times}
\usepackage{epsfig}
\usepackage{graphicx}
\usepackage{amsmath}
\usepackage{amssymb}
\usepackage{color, colortbl}
\usepackage{multirow}
\usepackage{makecell}
\usepackage{boldline}
\usepackage[accsupp]{axessibility}  

\newcommand\Mark[1]{\textsuperscript{#1}}
\DeclareMathOperator*{\argmin}{arg\,min}
\DeclareMathOperator*{\argmax}{arg\,max}
\wacvfinalcopy
\usepackage[pagebackref=true,breaklinks=true,colorlinks,bookmarks=false]{hyperref}

\begin{document}

\title{SporeAgent: Reinforced Scene-level Plausibility for Object Pose Refinement}
\author{
Dominik Bauer\Mark{1}, Timothy Patten\Mark{1,2} and Markus Vincze\Mark{1}\\
\begin{tabular}{cc}
\Mark{1}TU Wien, Austria & \Mark{2}University of Technology Sydney, Australia
\end{tabular}\\
{\tt\small \{bauer,patten,vincze\}@acin.tuwien.ac.at}
}
\maketitle

\thispagestyle{empty}

\begin{abstract}
Observational noise, inaccurate segmentation and ambiguity due to symmetry and occlusion lead to inaccurate object pose estimates. While depth- and RGB-based pose refinement approaches increase the accuracy of the resulting pose estimates, they are susceptible to ambiguity in the observation as they consider visual alignment.
We propose to leverage the fact that we often observe static, rigid scenes. Thus, the objects therein need to be under physically plausible poses. We show that considering plausibility reduces ambiguity and, in consequence, allows poses to be more accurately predicted in cluttered environments. To this end, we extend a recent RL-based registration approach towards iterative refinement of object poses.
Experiments on the LINEMOD and YCB-VIDEO datasets demonstrate the state-of-the-art performance of our depth-based refinement approach. 
Code is available at \href{https://www.github.com/dornik/sporeagent}{github.com/dornik/sporeagent}.
\end{abstract}

\section{Introduction}

\begin{figure}
    \centering
    \includegraphics[width=0.9\linewidth]{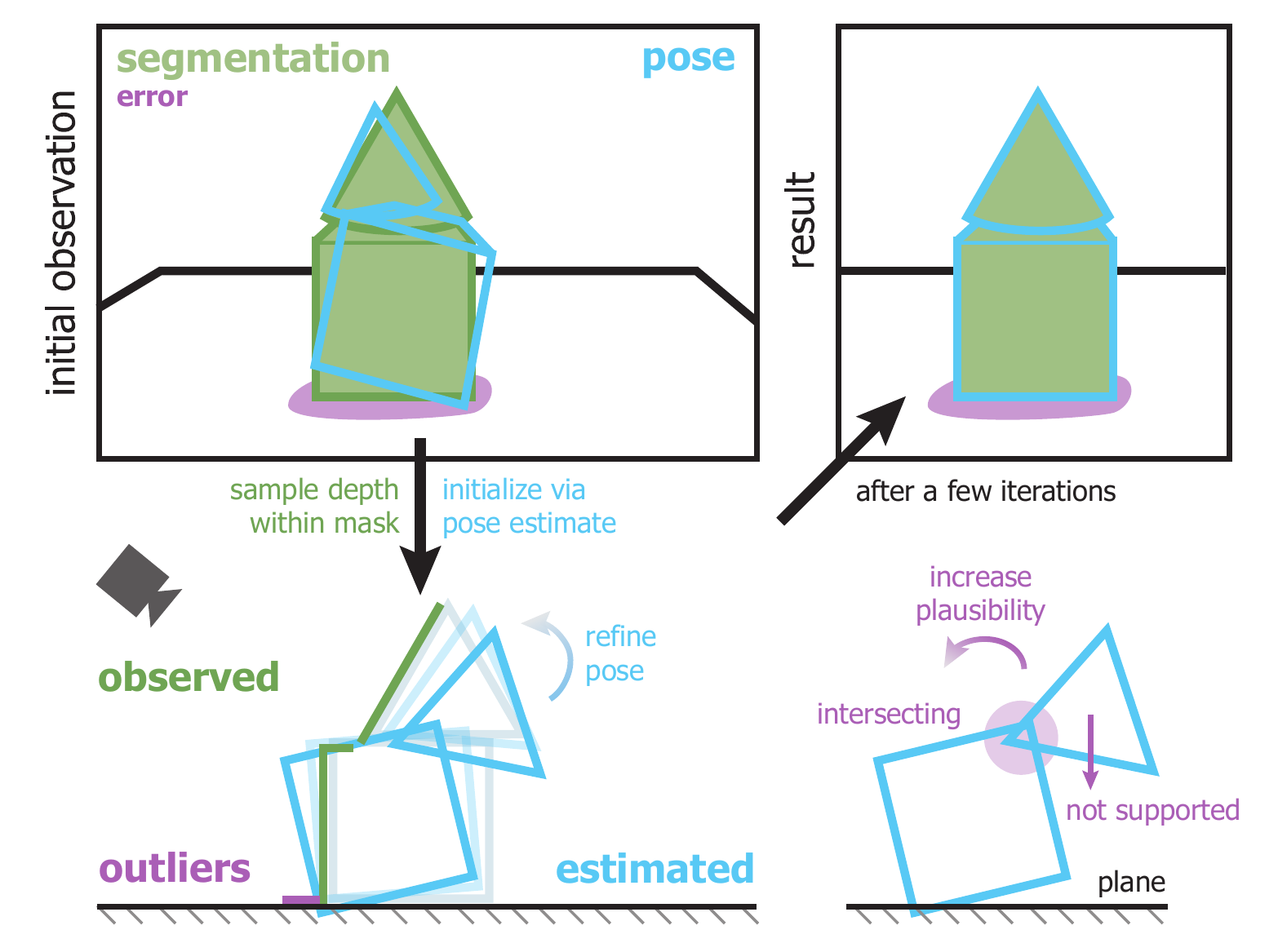}
    \caption{Objects are segmented and initial poses are estimated (upper left). The scene is initialized and all objects therein are refined iteratively. In each iteration, we more closely align the observed source and the target model point clouds  (lower left), while increasing physical plausibility of the scene (lower right).
    }
    \label{fig:teaser}
\end{figure}

Applications such as robotic grasping or augmented reality require the estimation of accurate object poses. This allows an observed scene to be represented by a set of 3D meshes and enables interaction with these scene objects. Given an initial pose estimate, the task of object pose refinement is to closely align an object's mesh with the corresponding image patch or point cloud to improve accuracy. 

Image-based approaches aim to align with an RGB(D) image patch or instance segmentation mask \cite{li2018deepim,zakharov2019dpod,shao2020pfrl,busam2020moveit}. Approaches using point clouds align the sampled surface of the object mesh with an observed point cloud that is computed from a depth image \cite{besl1992icp}. RGB images provide high resolution and textural information but distance and scale are ambiguous. Using absolute depth information resolves this ambiguity. However, depth sensors offer lower resolution and suffer from artifacts such as depth shadowing and quantization errors. While approaches combining both modalities achieve high accuracy \cite{li2018deepim,deng2021poserbpf,wang2019densefusion}, they are bound by the limitations of visual observability. \mbox{Indistinguishable} views of an object due to (self)occlusion and symmetry lead to ambiguous refinement targets.

To resolve such ambiguous situations, recent approaches additionally consider the physical configuration of the complete scene \cite{wada2020morefusion,mitash2018mcts,bauer2020verefine}. This additional source of information exploits the pose estimates of other scene objects, limiting the possible object interactions. These approaches are, however, limited to resolving collisions \cite{wada2020morefusion} or require physics simulation \cite{mitash2018mcts,bauer2020verefine}.

In contrast to prior work, we guide our refinement approach towards non-intersecting, non-floating and statically stable scenes by using a contact-based definition of physical plausibility \cite{bauer2020eccvw}. We extend a Reinforcement Learning-based point cloud registration approach \cite{bauer2021reagent} by considering surface distance as proxy for scene-level interaction information and reinforce a plausibility-based reward. Geometrical symmetry is considered to stabilize training using Imitiation and Reinforcement Learning (IL+RL).

Exploiting this additional source of information in combination with point cloud based alignment, our \textit{\textbf{S}cene-level \textbf{P}lausibility for \textbf{O}bject Pose \textbf{Re}finement} Agent (SporeAgent) achieves accuracy comparable to RGBD \mbox{approaches} and outperforms competing RGB and depth-only approaches. While RGB(D)-based methods need large amounts of training data to converge, we are able to use a fraction of training images, using only 1/100th of the real training images provided for YCB-VIDEO.

In summary, we achieve the presented results by:
\begin{itemize}
    \item integrating physical plausibility with IL+RL-based iterative refinement of object poses;
    \item simultaneously refining all objects in a scene, recovering from pose initialization errors due to ambiguous alignment using scene-level physical plausibility;
    \item considering geometrical symmetry, further dealing with indistinguishable views in the depth domain and
    \item achieving state-of-the-art accuracy, while requiring limited amounts of training data.
\end{itemize}
After an overview of previous work in Section \ref{sec:rw}, we briefly introduce the related tasks of point cloud registration and object pose refinement in Section \ref{sec:background}. Extensions to IL+RL-based point cloud registration tackling the discussed challenges in pose refinement are proposed in Section~\ref{sec:spore}. We evaluate our approach on the LINEMOD and YCB-VIDEO datasets and provide an ablation study of the proposed extensions in Section \ref{sec:experiments}. Section \ref{sec:conclusion} concludes the paper and gives an outlook to future work.

\section{Related Work}\label{sec:rw}
Closely related to depth-based pose refinement, the task of point cloud registration is recently approached as a learning task. Building on the core ideas of ICP \cite{besl1992icp}, learned feature representations improve the matching between corresponding points in the source and the target point clouds~\cite{wang2019dcp,choy2020dgr,yew2020rpmnet}. Yuan et al. \cite{yuan2020deepgmr} learn Gaussian Mixture Model parameters for probabilistic registration. A further line of work uses a learned global feature representation to represent the iterative registration state, combined with an application of the Lukas-Kanade algorithm \cite{aoki2019pointnetlk} or Reinforcement Learning (RL) \cite{bauer2021reagent}. Our refinement approach is based on the latter RL approach as it suited to incorporate domain knowledge such as physical plausibility.

Learning-based object pose refinement, in contrast, is commonly based on RGB images \cite{li2018deepim,shao2020pfrl,busam2020moveit,zakharov2019dpod} where the goal is to align a rendering of the object under estimated pose to an observed image patch. Seminal work by Li et al.~\cite{li2018deepim} uses optical flow features to predict the refinement transformation to more closely align the given image patches. The approaches of Shao et al. \cite{shao2020pfrl} and Busam et al.~\cite{busam2020moveit} consider this as a RL task and learn a policy that predicts discrete refinement actions. An alternative use of RL is proposed in \cite{krull2017poseagent}, where RL actions are selected as one of a pool of pose hypotheses to be refined in each iteration.

However, these approaches in point cloud registration and object pose refinement consider a single registration target (i.e., a single object) at a time. Especially in heavily cluttered scenes that may occur, for example, during robotic manipulation, consideration of the full scene of objects is shown to be beneficial. Labb{\'e} et al. \cite{labbe2020cosypose} propose a global refinement scheme, incorporating all scene objects in a joint pose optimization and incorporate multiple views of the scene. By fusing multiple views into a voxel grid, Wada et al. \cite{wada2020morefusion} additionally consider the collision between scene objects in their refinement method. A formalization of this concept of considering physical interactions between static scene objects for evaluation of pose estimates is proposed in \cite{bauer2020eccvw}, which is used for our definition of physical plausibility. In contrast, the methods proposed by Mitash et al. \cite{mitash2018mcts} and Bauer et al. \cite{bauer2020verefine} apply physics simulation to scene estimates, leveraging the simulated dynamics of the scene objects to improve pose hypotheses and verify the best aligned combinations with respect to the observed depth image. We employ a similar depth-based pose scoring to determine the best pose during refinement.

\section{Background: Point Cloud Registration using Imitation and Reinforcement Learning}\label{sec:background}
The task of depth-based object pose refinement is closely related to point cloud registration. In this section, we give a brief overview of registration, link it to refinement and introduce terms used throughout the paper. We additionally discuss how registration -- and by extension refinement -- is approached as a reinforcement learning task.

\subsection{Point Cloud Registration and Refinement} Assume we are given a source $X$ and target point cloud $Y$ that, for example, both represent a single object. The target is offset from the source by an unknown rigid transformation $T$ of form $[R \in SO(3), t \in \mathbb{R}^3]$. The task of registration is to find a transformation $\hat{T}$ such that the source and target are aligned again, i.e., $\hat{T} T^{-1} = I$. This transformation may be found directly or iteratively by updating the source with intermediary transformations $X_{i+1}=\hat{T}_i X$. Moreover, an initial estimate $\hat{T_0}$ may be provided. 

Translated to object pose refinement, the source is a partial and noise afflicted view of the object, represented by a point cloud, and the target is represented by a 3D mesh. The initial estimate is commonly computed using a dedicated object pose estimator.

\subsection{Reinforced Point Cloud Registration} Iteratively determining a transformation $\hat{T}_i$ that, with each iteration $i$, more closely aligns two point clouds $(X,Y)$ may be interpreted as taking a sequence of refinement actions. We will refer to such a sequence $\{a_1,...,a_{i+1}\}$ as a \textit{refinement trajectory}. Considering the prediction of such trajectories as a RL task, the goal is to determine a policy that selects a suitable refinement action $a_{i+1}$ given the current observation $O_i=(X_i,Y)$. In related work \cite{shao2020pfrl,bauer2021reagent}, the action space for this task is assumed to consist of a set of discrete steps per transformation dimension. Hence, in each iteration, the policy must determine a step per axis, separately for rotation and translation. Such an action, for example, might relate to taking a small translation step along x-axis. The policy for this action space is represented by a discrete probability distribution $\pi(a|O)$ that is conditioned on the observation.

Such a policy is predicted by the network architecture presented in \cite{bauer2021reagent}. A simplified PointNet \cite{qi2017pointnet} is used as embedding, consisting of three 1D-convolution layers of dimension $[64,256,1024]$. Source and target are processed independently with shared weights. The concatenation of the max pooling of the final layer constitutes the state vector and is used as representation of the observation. The state vector serves as input for the policy network, consisting of two separate heads for rotation and translation actions and an additional head for computing a value estimate for the current state. The value estimate serves as baseline for the advantage computation used in RL. All heads are implemented using fully-connected layers of dimension $[512,256,D]$ and the policy heads' output is interpreted as a discrete probability distribution over refinement actions.

An expert policy $\pi^*$ is able to determine the optimal actions in every iteration by taking the largest step towards the ground-truth pose $T$. Formally, such an expert policy may be defined \cite{bauer2021reagent} by taking the largest rotation and translation actions per-axis to minimize the residuals
\begin{equation}\label{eq:expert}
    \delta^R_i = R \hat{R_i}^\top, \quad \delta^t_i = t - \hat{t_i},
\end{equation}
where $T=[R,t]$ is the ground-truth transformation and $\hat{T_i}=[\hat{R_i},\hat{t_i}]$ is the current estimate. The actions predicted by this policy may be used in conjunction with IL approaches such as Behavioral Cloning, for which they serve as ground-truth labels during training.

A combination of IL with RL is proposed in \cite{bauer2021reagent}. An alignment-based reward $r_a$ is reinforced via Proximal Policy Optimization (PPO) \cite{schulman2017ppo} in a discounted task. For an action $a_{i+1}$ in iteration $i$, it is defined as
\begin{equation}\label{eq:alignment-reward}
    r_a = \begin{cases}
        -\rho^-,   &\quad CD(X_{i+1}, X^*) > CD(X_{i}, X^*)\\
        -\rho^0,   &\quad CD(X_{i+1}, X^*) = CD(X_{i}, X^*)\\
        \rho^+,    &\quad CD(X_{i+1}, X^*) < CD(X_{i}, X^*),\\
    \end{cases}
\end{equation}
where $CD$ is the Chamfer distance, $X^*=TX$ is the source under ground-truth pose and $(-\rho^-,-\rho^0,\rho^+)$ is a set of rewards for worsening, stagnating and improving the alignment with respect to the previous step. 
The loss from IL is blended with the PPO loss via a scaling factor $\alpha$ that affects the impact of PPO. We employ this combined approach as it is shown to quickly converge, while allowing further domain knowledge to be integrated via adaptation of the expert policy in Equation \eqref{eq:expert} and the reward in Equation \eqref{eq:alignment-reward}.

\section{Reinforced Scene-level Plausibility for Object Pose Refinement}\label{sec:spore}
In applications such as robotic grasping or augmented reality, erroneous object pose estimates may result in missed grasps or might disrupt the immersion of the user. To improve upon single-shot pose estimation, object pose refinement aims to iteratively increase the alignment of an object under estimated pose with an observation.

Object pose refinement from a single view further increases the difficulty of the task. First, objects are commonly observed in the context of complex scenes, requiring instance segmentation of the object of interest. Second, many classes of objects (e.g, cylindrical, box-shaped) feature geometrical symmetries, which makes alignment ambiguous. Finally, objects are only partially observable from a single view. This issue is aggravated in cluttered scenes due to occlusion between objects. 

To tackle each of these challenges, we extend the reinforced point cloud registration approach presented in Section \ref{sec:background} towards pose refinement. Consideration of geometrical symmetry in the expert policy and surface normals as additional input are discussed in Section \ref{sec:reg-to-ref}. As presented in Section \ref{sec:plausibility}, we furthermore integrate contact-based physical plausibility as input and at the RL stage to deal with visual ambiguity.

\begin{figure}
    \centering
    \includegraphics[width=\linewidth]{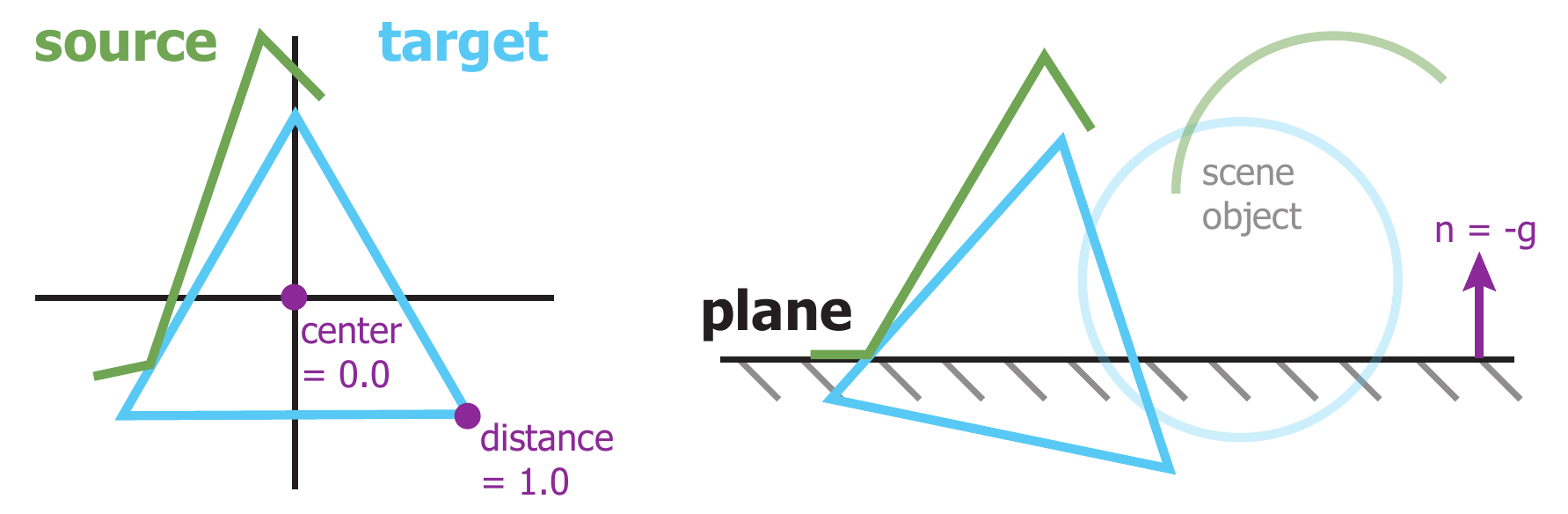}
    \caption{Representation for per-object refinement (left) and for computation of the scene-level information (right).}
    \label{fig:representation}
\end{figure}

\subsection{From Registration to Object Pose Refinement}\label{sec:reg-to-ref}
The baseline approach is designed for registration of noisy yet complete point clouds. In object pose refinement, however, the observed point cloud represents only a single partial view of the object. The required instance segmentation might include points belonging to the background or close-by objects. Hence, the noisy source only partially overlaps with the noise-free target that is sampled from the object's 3D mesh. The target may also have geometrical symmetries. The correct alignment between source and target point cloud is thus ambiguous, as any symmetrical pose will result in an indistinguishable observation of the object. The following extensions enable the registration to perform accurately on real data on the pose refinement task.

\subsubsection{Extending Input and Embedding}
To deal with inaccurate segmentation of the object, we adapt the input representation and extend it by further geometrical information. Surface normals are a strong geometrical cue for outlier points as the observed normals are subject to large changes at object borders. Source normals are estimated from the observed point cloud. For the target point cloud, they are retrieved from the mesh face from which the points are sampled.

\begin{figure}
    \centering
    \includegraphics[width=\linewidth]{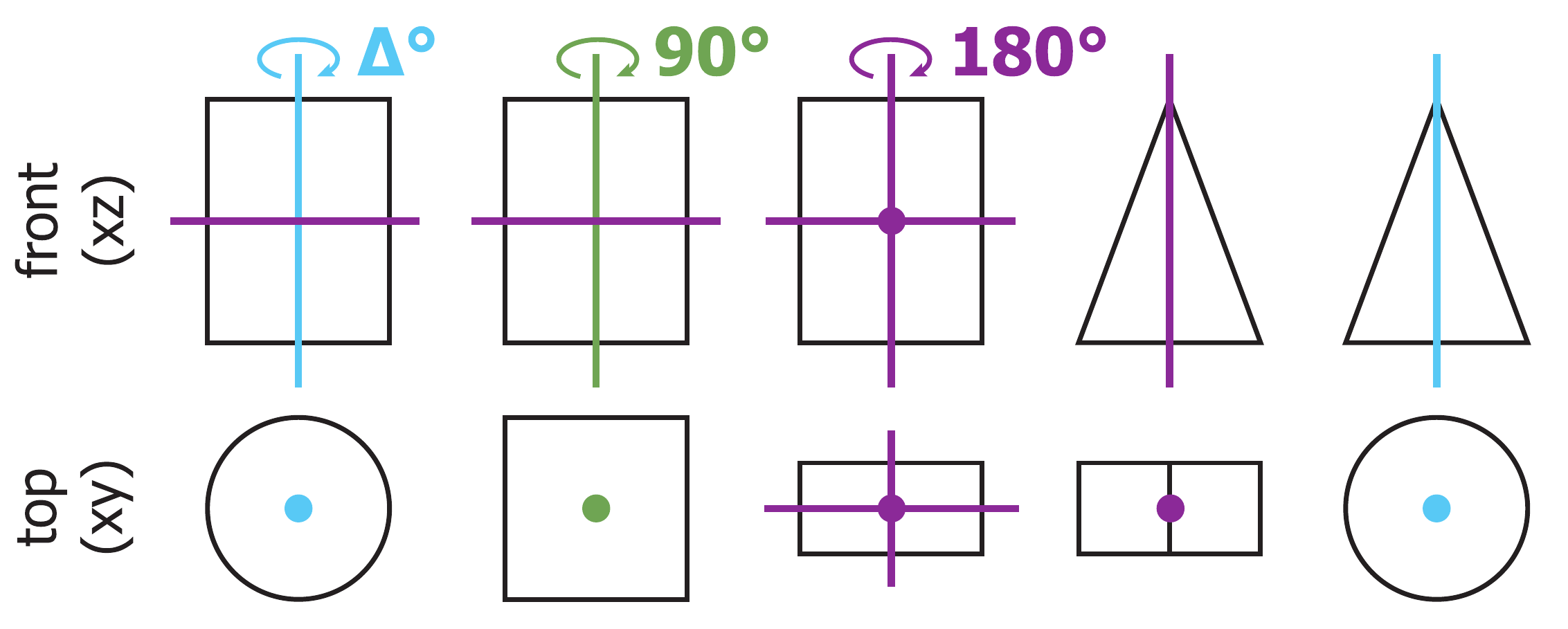}
    \caption{Geometrical symmetry classes. From left to right: Cylindrical, cuboid, box, front-back and rotational. Rotational symmetries (blue) are sampled with a resolution of $\Delta$deg.}
    \label{fig:symmetries}
\end{figure}

Normalizing the input point clouds provides an inductive bias that any point outside the unit sphere is either misaligned or an outlier. This also reduces the range of the potential inputs. Formally, the normalization with respect to the target $Y$ is defined by
\begin{equation}
    X' = (X - \mu_Y) / d_Y, \quad Y' = (Y - \mu_Y) / d_Y,
\end{equation}
where $\mu_Y$ is the centroid of the target and $d_Y$ is the maximal distance from the centroid to any point in the target. See Figure \ref{fig:representation} (left) for an illustration.
Supporting the uniform coverage of the input space further, we align objects into a canonical orientation in which the major symmetry axis is aligned with the z-axis. Cylindrical, cuboid and box-shaped objects are oriented uniformly as shown in Figure~\ref{fig:symmetries}. Thereby, objects' targets will more closely align and thus leverage similar geometrical features. This supports training and results in higher refinement accuracy as discussed in the ablation study in Section \ref{sec:ablation}.

To accommodate this additional input and further extensions to be discussed in the following sections, we add two additional layers to the baseline's embedding network, resulting in five 1D-convolution layers of dimension $[64,128,256,512,1024]$ in total.

Moreover, outlier points are pruned using a segmentation branch consisting of four 1D-convolution layers of dimension $[512,256,128,1]$. The concatenation of the first and last layer's feature embedding as well as a one-hot class vector results in a $1088+k$ dimensional input, where $k$ is the number of classes. Points that are labeled \textit{outliers} are ignored in the max pooling operation and thus do not contribute towards the state vector. This outlier removal is trained using the ground-truth instance labels and a binary cross-entropy loss, scaled by a factor $\beta$.

\subsubsection{Symmetry-aware Expert Policy}
The expert policy presented in Section \ref{sec:background} provides the goal trajectories for IL. It is defined to minimize the residual between the estimated and ground-truth pose. However, for symmetrical objects, the object under ground-truth pose is indistinguishable from a transformation by an alternative symmetrical pose. The expert labels are thus inconsistent with respect to the observation and the agent will receive a varying loss for seemingly equivalent actions.

To address this issue, we consider multiple equivalent ground-truth poses of form $[R_s, t_s]$ for symmetrical objects. The expert should follow the shortest trajectory and thus move towards the symmetrical pose that is closest to the current estimate $\hat{T_i}$. Due to the symmetry axes coinciding with the origin in our canonical representation of the objects, the symmetry transformations do not involve any translation. The index $s_i$ of the closest pose is thus determined by taking the residual rotation with the smallest angle
\begin{equation}
\begin{split}
    s_i &= \argmin_{s} \arccos \frac{trace(R_s \hat{R_i}^\top) - 1}{2} \\
        &= \argmax_{s} trace(R_s \hat{R_i}^\top).
\end{split}
\end{equation}
The expert actions are computed as before by retrieving the symmetry-aware residuals using Equation \eqref{eq:expert} with respect to $[R_{s_i}, t_{s_i}]$ as ground truth.

Since most related approaches exploit RGB information, the available symmetry annotations for common pose estimation datasets \cite{hodan2020bop} consider the texture of objects as well as their geometry. Hence, we need to annotate additional geometrical symmetries for use with depth-only observations. In canonical representation, this reduces to assigning objects to one of the five symmetry classes, as illustrated in Figure \ref{fig:symmetries}, in our experiments\footnote{The transformations to our canonical representation and the corresponding symmetry annotations per object are provided with our code.}.

\begin{figure*}
    \centering
    \includegraphics[width=0.9\linewidth]{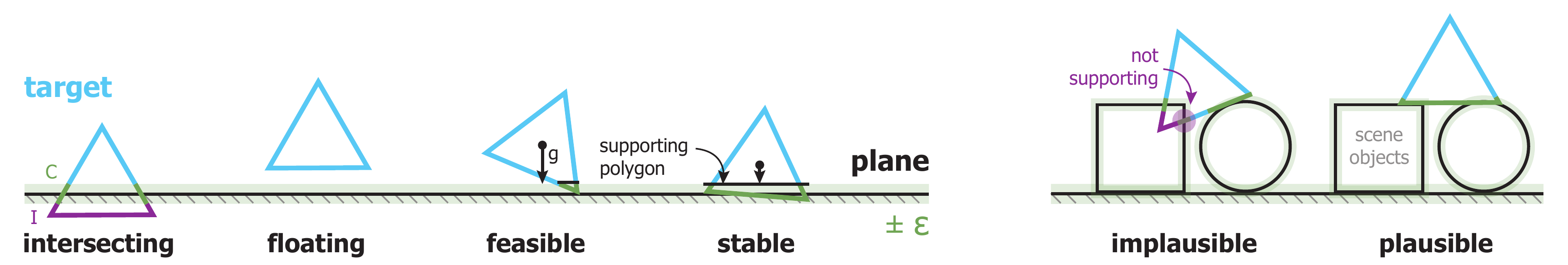}
    \caption{Definition of physical plausibility based on critical points for a single object (left) and a scene (right). If feasible, the CoM projected in gravity direction must intersect the support polygon (convex hull of supported points) to be considered stable.}
    \label{fig:plausibility}
\end{figure*}

\subsection{Scene-level Physical Plausibility}\label{sec:plausibility}
Visual information alone might still be ambiguous due to occlusion or observational noise. However, assuming the observed scene is static, physically plausible interactions between an object and its support limit the space of admissible object poses within the scene.

\subsubsection{Definition of Physical Plausibility}
Individual aspects of physical plausibility such as collision are modeled using voxel grids \cite{wada2020morefusion} or by the dynamics computed using physics simulation \cite{mitash2018mcts,bauer2020verefine}. In contrast, physical plausibility defined in \cite{bauer2020eccvw} is based on the interacting points between objects. As such, this approach is well suited to be incorporated with point cloud inputs.

As illustrated in Figure \ref{fig:plausibility} (left), the signed distance between two objects serves as a basis for the definition of points that are critical for physical plausibility. The signed distance is the Euclidean distance to the closest surface point, with points inside the surface having negative sign. We refer to it as \textit{surface distance} $d(x)$ of a point $x$. 

Points farther within another object than a threshold $\varepsilon$ are considered \textit{intersecting}, as defined by
\begin{equation}\label{eq:intersecting}
    I = \{x \in X: d(x) < -\varepsilon\}.
\end{equation}
Points within an absolute distance of $\varepsilon$ are considered \textit{in contact}, or formally
\begin{equation}\label{eq:contact}
    C = \{x \in X: | d(x)| < \varepsilon\}.
\end{equation}
In addition to the definitions of \cite{bauer2020eccvw}, we consider a contact to be \textit{supported}, if
\begin{equation}\label{eq:supported}
    S = \{c \in C: \cos(n_y(c) \cdot g) < 0\},
\end{equation}
where $n_y(c)$ is the normal of the closest point $y$ to $c$ in the scene and $g$ the gravity direction. These critical points allow to define the plausibility conditions, illustrated in Figure \ref{fig:plausibility}. 

If there is at least one intersecting point, the object is considered \textit{intersecting}. If an object has no contact point, it is consider to be \textit{floating}. If it is neither intersecting nor floating, the object is defined to be under a \textit{feasible} pose within the current scene. Finally, if the center of mass (CoM) of the object -- projected in the gravity direction -- falls within the supporting polygon, the object is \textit{stable}. The supporting polygon is defined as the convex hull of the supported points $S$. An object that is both feasible and stable under a given pose is considered to be \textit{physically plausible}.
\subsubsection{Computing Surface Distance and Critical Points}
As the used definition of physical plausibility depends on the surface distance, we need to efficiently compute it for the whole scene in every refinement iteration. Moreover, we need to determine an initial supporting plane and gravity direction from the observation.

Assuming the supporting plane is sufficiently visible in the scene, we use a RANSAC-based approach to fit a plane to the subset of the source point cloud that is assigned a background label in the instance segmentation. The plane's pose with respect to the camera is then given by $\hat{P}$. To determine the surface distance between scene objects, we transform all corresponding object point clouds to this plane. The sources are already in camera space and may thus be transformed by $\hat{P}$. For the targets, we consider the current pose estimate $\hat{T_i}$ and transform targets by $\hat{P}\hat{T_i}^{-1}$.

The surface distance with respect to the plane is now trivially obtained by the z-coordinate of the queried point~$x$. For the distance between objects, the nearest neighbor in each scene object is computed per queried point. The point lies within the corresponding scene object if
\begin{equation}
    \cos (n_y(x) \cdot (y - x) ) > 0,
\end{equation}
that is, if the normal of the nearest neighbor $n_y(x)$ points in the same direction as the vector from $x$ to the nearest neighbor $y$. To make this test more robust, we compute it for the $k$ nearest neighbors and consider the point to lie inside if it holds for a certain fraction of $k$. Note that a similar quorum check is used to compute the supported points $S$. 

The final surface distance per point $x$ is computed by taking the minimum over the distances to all scene objects (and the plane). The critical points of the queried point cloud are then computed using Equations \eqref{eq:intersecting}--\eqref{eq:supported}, with the gravity direction given by the normal of the supporting plane. 

Initially, the plausibility information derived from the computed surface distances is inaccurate, since the initial poses of the scene objects themselves are inaccurate. We therefore consider all object poses in parallel -- as all poses improve, so does the scene-level plausibility information. 

\subsubsection{Leveraging Plausibility for Refinement}
We want to exploit the additional source of information and guide the agent towards plausible poses. The most straightforward approach is to add the surface distance as additional input. Intuitively, the agent should move away from points that have a negative surface distance (intersecting) and keep some points close to zero distance (in contact). As the surface distance is available for both source and target, this also situates the object within the scene -- if we assume two points to correspond, their surface distance under the estimated pose should also correspond.

To enforce this behavior and to also provide a training signal for the full trajectory, we add a plausibility-based reward term. To reinforce actions that lead to poses under which the object rest stably within the scene, we define
\begin{equation}
    r_p = \begin{cases}
    +\rho_p, \text{\quad if stable}, \\
    -\rho_p, \text{\quad otherwise}.
    \end{cases}
\end{equation}
Note that a necessary condition for stability is feasibility. Thus, by reinforcing stability, feasibility is implicitly considered too. This reward is combined with the alignment-based reward in Equation~\eqref{eq:alignment-reward} in a discounted task. Hence, if actions lead to a stable pose later on, the reward also reinforces these earlier actions via the discounted return.

\subsubsection{Rendering-based Pose Scoring}
Due to imprecise segmentation, the refinement may consider close-by objects or points sampled from the background. Consequently, the alignment after the final iteration might be lower than in an intermediary step. This is all the more important as poses of scene objects influence one another. Related work \cite{deng2021poserbpf,bauer2020verefine} deals with this issue by rendering-based verification of pose hypotheses.

In each iteration, the object under currently estimated pose is rendered, giving a depth image $\hat{I}_d$ and a normal image $\hat{I}_n$. These are compared to the observed depth and normal images $I_d$ and $I_n$, masked by the estimated segmentation for the corresponding object. Based on \cite{bauer2020verefine}, we define the per-pixel score $e(p)$ by
\begin{equation}
    e(p) = (e_d + e_n)/2, \quad \forall p \in \hat{I}_d > 0 \cup I_d > 0,
\end{equation}
where the depth and the normal scores $e_d,e_n$ are given by
\begin{equation}
    \begin{split}
        e_d(p) = 1 - \min(1, |\hat{I}_d(p) - I_d(p)| / \tau_d), \\
        e_n(p) = 1 - \min(1, (1 - cos \hat{I}_n(p) \cdot I_n(p)) / \tau_n), \\
    \end{split}
\end{equation}
with thresholds $\tau_d$ and $\tau_n$ clamping and scaling the respective error to the range $[0, 1]$. Note that for the normal error $e_n$, the cosine is clamped to $[0, 1]$. The mean over the per-pixel scores is used to score the corresponding pose and the one with maximal score is returned as refinement result.

\section{Experiments}\label{sec:experiments}
\begin{figure}
    \centering
    \includegraphics[trim=300 374 300 74, clip, width=0.5\linewidth]{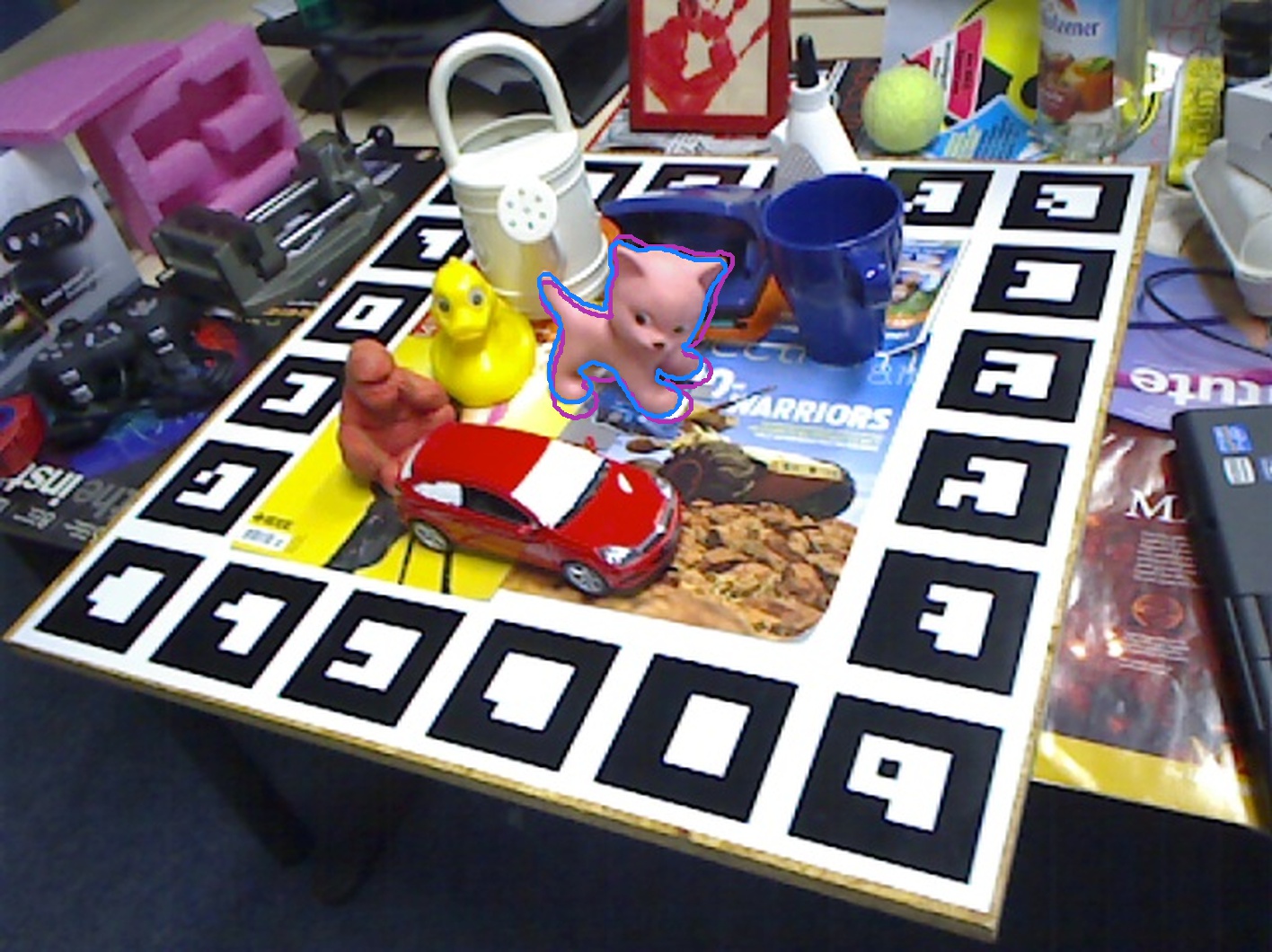}\includegraphics[width=0.5\linewidth]{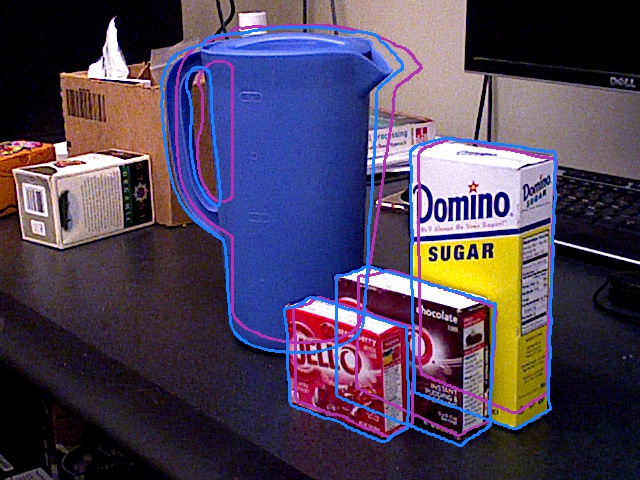}
    \caption{Initial (purple) and final pose (blue) on LINEMOD (left, cropped) and YCB-VIDEO (right). Best viewed digitally.}
    \label{fig:qualitative}
\end{figure}

We evaluate the proposed depth-based pose refinement approach in comparison with state-of-the-art methods, provide an ablation study for the proposed extensions and discuss failure cases. Qualitative examples are shown in Figure \ref{fig:qualitative}.

\textbf{Datasets:} The single object scenario is evaluated on the LINEMOD dataset (LM) \cite{hinterstoisser2012adi}, consisting of 15 test scenes showing one object in a cluttered environment each but with only minor occlusion of the target object. The test split defined in related work \cite{brachmann2016uncertainty,rad2017bb8,tekin2018real} is used, omitting scenes 3 and 7. The YCB-VIDEO dataset (YCBV) \cite{xiang2017posecnn} features more complex scene-level interactions and heavy occlusion with test scenes consisting of three to six YCB objects \cite{calli2015ycb}. In contrast to most competing approaches, we do not use any additional synthetic training data. Moreover, on YCBV, we only use 1/100th of the real training images.

\textbf{Metrics:} Evaluation metrics used in related work are the Average Distance of Model Points (ADD) and ADD with Indistinguishable Views (ADI)~\cite{hinterstoisser2012adi}. The ADD is defined as the mean distance between corresponding model points $m \in M$ under estimated pose $\hat{T}$ and under ground-truth pose $T$. To deal with symmetrical objects, the ADI computes the mean distance between the nearest neighbors under either pose. Formally, the metrics are defined as
\begin{equation}
\begin{split}
    ADD &= \frac{1}{|M|} \sum_{m \in M} ||\hat{T}m - Tm||_2,\\
    ADI &= \frac{1}{|M|} \sum_{m_1 \in M} \min_{m_2 \in M} ||\hat{T}m_1 - Tm_2||_2.
\end{split}
\end{equation}
We abbreviate a mixed usage of both metrics with \textit{AD}. For AD, objects considered symmetrical are evaluated using the ADI metric and using ADD otherwise. Note that the distinction between (non)symmetrical objects in related work also considers textural information in the color image which is not observable by our depth-based approach. %
These metrics are computed for all $N$ test samples and the recall for a given precision threshold $th$ is defined as
\begin{equation}
    AD_{th} = \frac{1}{N} \sum_{i=1}^N \begin{cases}
       0, & AD_i > th\\
       1, & AD_i \leq th.
    \end{cases}
\end{equation}
For the \textit{Area Under the precision-recall Curve} (AUC), the precision threshold $th$ is varied within $[0,0.1d]$ and the area under resulting curve of recall values is reported.

\textbf{Baselines:} We compare our approach to state-of-the art pose refinement methods that use RGB (DeepIM \cite{li2018deepim}, PoseRBPF \cite{deng2021poserbpf}, PFRL \cite{shao2020pfrl}), depth (Point-to-Plane ICP (P2Pl-ICP) \cite{chen1992p2pl,open3d}, Iterative Collision Check with ICP (ICC-ICP) \cite{wada2020morefusion}, VeREFINE \cite{bauer2020verefine}, the ICP-based multi-hypothesis approach (Multi-ICP) in \cite{xiang2017posecnn}) or a combination of both modalities (RGBD versions of DeepIM \cite{li2018deepim} and PoseRBPF \cite{deng2021poserbpf}). The RGB-based method of Shao et al. \cite{shao2020pfrl} (PFRL) is conceptionally close to our approach as it leverages RL to refine the pose by aligning the observed mask. In terms of used modality, the ICP-based approaches are considered for direct comparison. Note that the initialization by PoseCNN \cite{xiang2017posecnn} only provides a single pose hypothesis; we indicate the use of additional pose hypotheses by \textit{Multi-ICP} with $^*$. P2Pl-ICP and VeREFINE (using P2Pl-ICP) are given a budget of 30 refinement iterations. The correspondence threshold for P2Pl-ICP, given as index in the results, is defined as fraction of the object size. By \textit{(RGB)D} we indicate that a method uses depth for refinement but is initialized by a method that additionally uses color information.

\textbf{Implementation Details:} 
Following related work \cite{shao2020pfrl,li2018deepim,deng2021poserbpf,labbe2020cosypose}, we use the instance segmentation masks and poses estimated by PoseCNN \cite{xiang2017posecnn} for initialization.

For the refinement actions, we use step sizes of $[0.0033, 0.01, 0.03, 0.09, 0.27]$ in positive and negative direction plus a ``stop'' action with step size $0$. Step sizes are interpreted in radians for rotation and in units of the normalized representation for translation. Rotational symmetries are sampled with a resolution of $5$deg. The threshold~$\varepsilon$ for computing the critical points for plausibility is experimentally determined with $1cm$. The reward function uses $\rho=(-0.6,-0.1,0.5)$ for alignment and $\rho_p=0.5$ for plausibility. The RL loss term is scaled by $\alpha=0.1$ on LM and by $0.2$ on YCBV. The outlier-removal loss term is scaled by $\beta=7$. The remaining network parameters are chosen as in \cite{bauer2021reagent}. The segmentation masks are provided as a single channel image, missing information where they overlap, and the clamps in YCBV are confused in many frames. To compute the verification score, we thus merge both masks and use thresholds $\tau_d=2cm$ and $\tau_n=0.7$.

During training, we generate a replay buffer of 128 object trajectories each, delaying network updates until it is filled and sampling batches from the shuffled buffer. The training samples are augmented by an artificial segmentation error, random pose error for initialization and a random error in the pose of the plane. The segmentation error selects a random pixel within the ground-truth mask. The nearest neighbors to this pixel are determined and $p$\% of the foreground and $100-p$\% of the background neighbors are sampled. This simulates occlusion and inaccurate segmentation, including parts of the background. Error magnitudes are given with the experiments on the respective dataset. We train SporeAgent for 100 epochs, starting with a learning rate of $10^{-3}$ and halving it every 20 epochs.

\subsection{Single Object: LINEMOD}
For training, we uniformly randomly choose the foreground fraction $p\in[50,100\%]$. The initial pose is sampled by rotation about a uniformly random rotation vector by a random magnitude in $[0, 90deg]$. Similarly, the translation is in direction of a uniformly random vector of magnitude $[0.0,1.0]$ units in the normalized space of the respective object. Additionally, the estimated plane is jittered by a random rotation in $[0,5deg]$ and a random translation in $[0,2cm]$, sampled as for to the initial pose.

\begin{table}[]
\footnotesize
\setlength\tabcolsep{1.3ex}
    \centering
    \begin{tabular}{l|ccc|l}
         & \makecell{AD ($\uparrow$)\\$<0.10d$} & \makecell{AD ($\uparrow$)\\$<0.05d$} & \makecell{AD ($\uparrow$)\\$<0.02d$} & Modality \\\hline
\rowcolor[rgb]{0.95,0.95,0.95}
PoseCNN \cite{xiang2017posecnn}             & 62.7 & 26.9 &  3.3 & RGB \\
PFRL \cite{shao2020pfrl}                & 79.7 & -- & -- & RGB \\
DeepIM \cite{li2018deepim}             & 88.6 & 69.2 & 30.9 & RGB \\ \hline
P2Pl-ICP$_{0.2}$ \cite{open3d}      & 90.6 & 81.2 & 36.8 & (RGB)D \\
VeREFINE$_{0.2}$ \cite{bauer2020verefine} & 95.4 & 87.5 & \textit{39.5} & (RGB)D \\
P2Pl-ICP$_{0.3}$ \cite{chen1992p2pl,open3d}      & 92.6 & 79.8 & 29.9 & (RGB)D \\
VeREFINE$_{0.3}$ \cite{bauer2020verefine} & \textit{96.1} & 85.8 & 32.5 & (RGB)D \\
Multi-ICP$^*$ \cite{xiang2017posecnn}      & \textbf{99.3} & \textit{89.9} & 35.6 & (RGB)D \\
SporeAgent (ours)   & \textbf{99.3} & \textbf{93.7} & \textbf{50.3} & (RGB)D \\
    \end{tabular}\vspace{1ex}
    \caption{Results on LINEMOD (mean over per-class results).
    }
    \label{tab:linemod}
\end{table}

Table \ref{tab:linemod} shows the mean class recalls for varying precision thresholds with respect to the object diameter $d$. As shown, SporeAgent outperforms related approaches by a large margin of 3.8\% using the $0.05d$ threshold and 10.8\% using the $0.02d$ as compared to the next best methods.

\subsection{Full Scene: YCB-VIDEO}
As YCBV already features large amounts of occlusion, we increase the foreground fraction $p$ to the range $[80,100\%]$ during training. In addition to this, the training samples are more challenging and we thus reduce the initial pose error to $[0,75deg]$ and $[0,0.75]$ translation units. The same plane error as on LM is used.

\begin{table}[]
\footnotesize
\setlength\tabcolsep{1.65ex}
    \centering
    \begin{tabular}{l|ccc|l}
         & \makecell{ADD ($\uparrow$)\\AUC} & \makecell{AD ($\uparrow$)\\AUC} & \makecell{ADI ($\uparrow$)\\AUC} & Modality \\\hline
\rowcolor[rgb]{0.95,0.95,0.95}
PoseCNN \cite{xiang2017posecnn}       & 51.5 & 61.3 & 75.2 & RGB \\
CosyPose \cite{labbe2020cosypose}      & --   & 84.5 & 89.5 & RGB \\
PoseRBPF \cite{deng2021poserbpf}      & 59.9 & --   & 77.5 & RGB \\
DeepIM \cite{li2018deepim}        & 71.7 & 81.9 & 88.1 & RGB \\ \hline
PoseRBPF \cite{deng2021poserbpf}      & \textbf{80.8} & 88.5 & 93.3 & RGBD \\
DeepIM \cite{li2018deepim}        & \textit{80.7} & \textbf{90.4} & \textbf{94.0} & RGBD \\ \hline
ICC-ICP \cite{wada2020morefusion}       & 67.5 & 77.0 & 85.6 & (RGB)D \\
P2Pl-ICP$_{1.0}$ \cite{chen1992p2pl,open3d}      & 68.2 & 79.2 & 87.8 & (RGB)D \\
VeREFINE$_{1.0}$ \cite{bauer2020verefine} & 70.1 & 81.0 & 88.8 & (RGB)D \\
Multi-ICP$^*$ \cite{xiang2017posecnn}     & 77.4 & 86.6 & 92.6 & (RGB)D \\
SporeAgent (ours)   & 79.0 & \textit{88.8} & \textit{93.6} & (RGB)D \\
    \end{tabular}\vspace{1ex}
    \caption{Results on YCB-VIDEO (mean over per-class results).
    }
    \label{tab:ycbv}
\end{table}

The results in Table \ref{tab:ycbv} report the Area Under the precision-recall Curve (AUC) for a varying precision threshold of $[0,10cm]$, averaged over per-class results. On the ADI metric, our approach achieves accuracy on par with RGBD-based methods, being only second to the RGBD version of DeepIM \cite{li2018deepim}. Although our approach uses depth information and, as such, may not consider textural symmetries reflected by high ADD and AD scores, we are able to achieve higher accuracy than competing RGB-based approaches. Compared to the best other evaluated depth-based approach (Multi-ICP \cite{xiang2017posecnn}), we improve accuracy by 1 to 2.2\%. Over the best single-hypothesis depth-based approach (VeREFINE \cite{bauer2020verefine}), the improvement is even higher with 4.8 to 8.9\%. Moreover, note that compared to the best performing learning-based approaches, we require orders of magnitude less real and no synthetic training data.

We would like to emphasize that the results reported for PoseCNN are recomputed from the poses provided by the authors of \cite{xiang2017posecnn}, as those poses serve as initialization to our method. They however differ slightly from the scores reported in their paper. Results for Multi-ICP are also computed from the poses provided by the authors of \cite{xiang2017posecnn,li2018deepim}.

\subsection{Ablation Study}\label{sec:ablation}
To evaluate the impact of each of the proposed parts of our method, we train separate models on LM and YCBV with each part removed individually.

\begin{table}[]
\setlength\tabcolsep{1.0ex}
\footnotesize
    \centering
    \begin{tabular}{l|ccc}
                    & \makecell{AD ($\uparrow$)\\$<0.10d$} & \makecell{AD ($\uparrow$)\\$<0.05d$} & \makecell{AD ($\uparrow$)\\$<0.02d$}\\\hline
\rowcolor[rgb]{0.95,0.95,0.95}
SporeAgent           & 99.3 & 93.7 & 50.3 \\
no normals           & 99.1 & 93.5 & \textbf{50.8} \\
no outlier removal      & 99.3 & 93.6 & 50.1 \\
no stability reward  & 99.3 & 93.7 & 50.0 \\
no surface distance  & \textbf{99.4} & \textbf{94.0} & 46.3 \\
no pose scoring      & 98.9 & 92.5 & 46.9 \\ \hline
\rowcolor[rgb]{0.95,0.95,0.95}
Baseline (IL)       & 98.8 & 90.6 & 39.7 \\
    \end{tabular}\vspace{1ex}
    \caption{Ablation on LINEMOD.}
    \label{tab:ablation_lm}
\end{table}

Table \ref{tab:ablation_lm} shows results on LM. We see that the scene-level information is essential to achieve high recall under restrictive thresholds. Moreover, the rendering-based scoring is able to determine the best-aligned intermediary pose -- without the scoring, the last iterations jitter between tight alignment and observational noise.

\begin{table}[]
\setlength\tabcolsep{1.0ex}
\footnotesize
    \centering
    \begin{tabular}{l|ccc}
                    & \makecell{ADD ($\uparrow$)\\AUC} & \makecell{AD ($\uparrow$)\\AUC} & \makecell{ADI ($\uparrow$)\\AUC}\\\hline
\rowcolor[rgb]{0.95,0.95,0.95}
SporeAgent           & \textbf{79.0} & \textbf{88.8} & \textbf{93.6} \\
no canonical         & 76.4 & 86.7 & 92.6 \\
no symmetry          & 78.2 & 88.0 & 93.4 \\
no outlier removal      & 78.6 & 88.5 & 93.5 \\ 
no stability reward  & 78.2 & 88.1 & 93.4 \\
no surface distance  & 77.3 & 87.3 & 92.9 \\
no pose scoring      & 78.3 & 88.1 & 93.0 \\ \hline
\rowcolor[rgb]{0.95,0.95,0.95}
Baseline (IL) + normals  & 72.2 & 82.2 & 90.7 \\
    \end{tabular}\vspace{1ex}
    \caption{Ablation on YCB-VIDEO.
    }
    \label{tab:ablation_ycbv}
\end{table}

The ablation study on YCBV, shown in Table \ref{tab:ablation_ycbv}, highlights the benefit of representing the target point clouds in a canonical frame. As on LM, the consideration of the physical plausibility improves accuracy. The scene-level information is able to support the refinement, even when computed from initially inaccurate object poses.

\begin{figure}
    \centering
    \includegraphics[width=\linewidth]{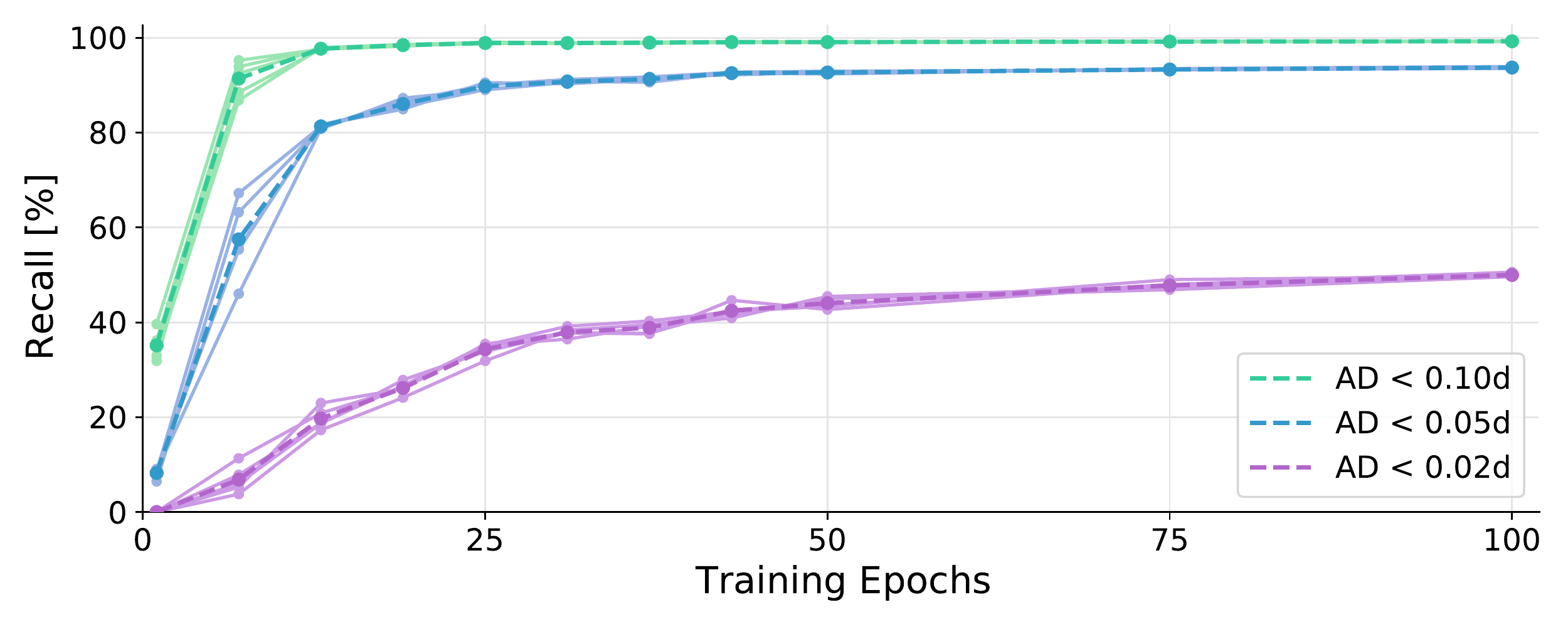}
    \caption{Convergence of the mean recall per epoch (dashed) on LINEMOD for 5 random seeds (solid). Best viewed digitally.}
    \label{fig:convergence}
\end{figure}

\textbf{Convergence with Varying Random Seed: }Figure~\ref{fig:convergence} shows the training convergence of SporeAgent on LM for 5 different random seeds. The reported AD recalls are obtained by evaluation on the test set using the current weights after the corresponding training epoch. After 50 epochs, the final performance for the $0.10d$ threshold is already achieved and the recall is within $1\%$ for the $0.05d$ threshold. The standard deviation for the last epoch is $0.02\%, 0.2\%$ and $0.3\%$ for the $0.10d, 0.05d$ and $0.02d$ threshold, respectively.

\begin{table}[]
\setlength\tabcolsep{0.3ex}
\footnotesize
    \centering
    \begin{tabular}{l|ccc}
     fraction & \makecell{ADD ($\uparrow$)\\AUC} & \makecell{AD ($\uparrow$)\\AUC} & \makecell{ADI ($\uparrow$)\\AUC}\\\hline
1/400th              & 75.4 & 86.2 & 92.5 \\
1/200th              & 77.9 & 87.9 & 93.3 \\
\rowcolor[rgb]{0.95,0.95,0.95}
1/100th              & \textbf{79.0} & \textbf{88.8} & \textbf{93.6} \\
    \end{tabular}\hspace{2.5ex}
        \begin{tabular}{l|ccc}
points  & \makecell{ADD ($\uparrow$)\\AUC} & \makecell{AD ($\uparrow$)\\AUC} & \makecell{ADI ($\uparrow$)\\AUC}\\\hline
256              & 77.9 & 87.8 & 93.3 \\
512              & 78.6 & 88.4 & 93.5 \\
\rowcolor[rgb]{0.95,0.95,0.95}
1024              & \textbf{79.0} & \textbf{88.8} & \textbf{93.6} \\
    \end{tabular}
    
    \vspace{1ex}
    \caption{Results by fraction of training split used (left) and number of points sampled for refinement (right) on YCB-VIDEO.}
    \label{tab:ablation_data}
\end{table}

\textbf{Number of Samples: }We evaluate the impact of the number of training samples and point samples used on YCBV. As shown in Table \ref{tab:ablation_data} (left), the performance of Multi-ICP \cite{xiang2017posecnn} on the AD AUC and ADI AUC metrics is achieved already using 1/400th of the available real training images. The number of points may be reduced to improve runtime and memory footprint while maintaining high accuracy, indicated by the results in Table \ref{tab:ablation_data} (right).

\begin{table}[]
\footnotesize
    \centering
    \begin{tabular}{l|ccc}
         & \makecell{ADD AUC ($\uparrow$)} & \makecell{AD AUC ($\uparrow$)} & \makecell{ADI AUC ($\uparrow$)} \\\hline
\rowcolor[rgb]{0.95,0.95,0.95}
init (PoseCNN) & 51.5 & 61.3 & 75.2 \\
1 iteration & 64.4 & 74.6 & 84.5 \\
2           & 71.0 & 81.0 & 88.5 \\
\rowcolor[rgb]{0.95,0.95,0.95}
Multi-ICP$^*$ \cite{xiang2017posecnn}   & 77.4 & 86.6 & 92.6\\
5           & 77.9 & 87.6 & 92.9 \\
10          & \textbf{79.0} & \textbf{88.8} & \textbf{93.6} \\
    \end{tabular}\vspace{1ex}
    \caption{Influence of number of iterations on results of SporeAgent on YCB-VIDEO. The results of PoseCNN (our initialization) and the ICP-based multi-hypothesis approach in \cite{xiang2017posecnn} are shown (gray).}
    \label{tab:iterations}
\end{table}

\begin{figure}
    \centering
    \includegraphics[width=\linewidth]{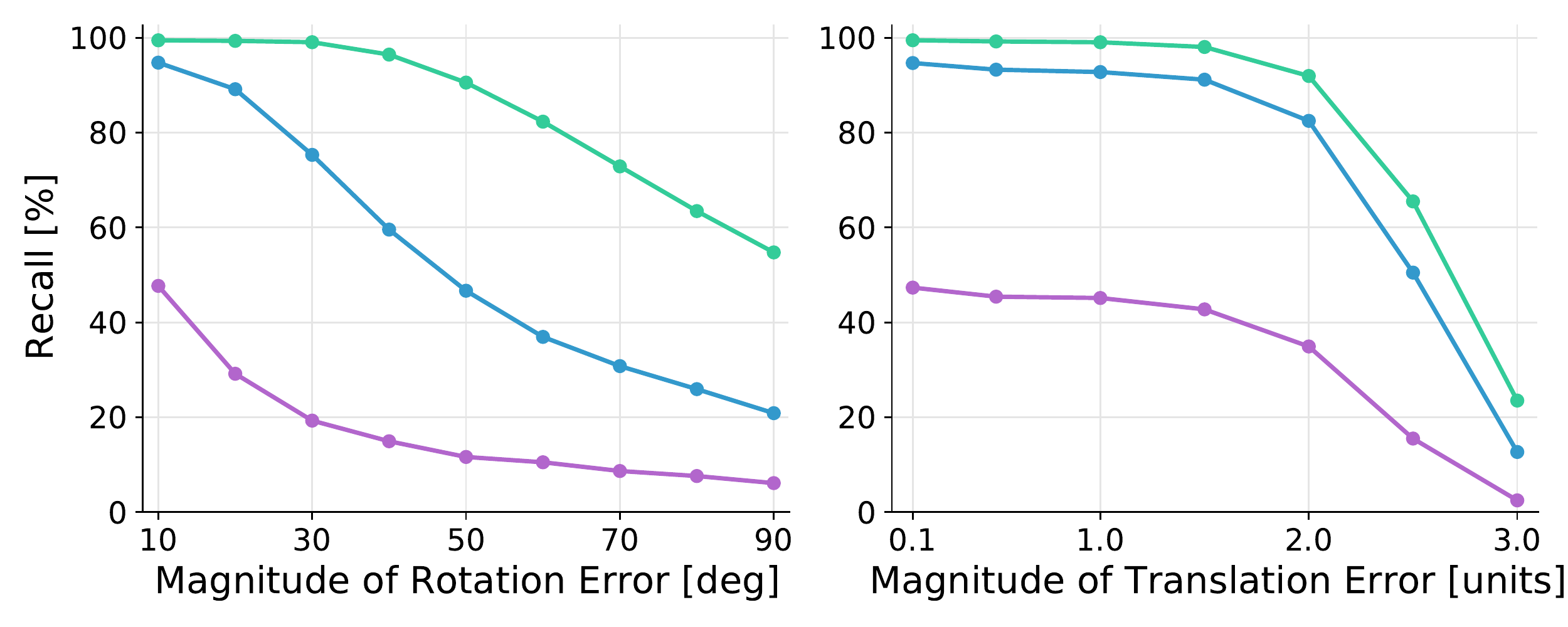}
    \caption{AD recalls on LINEMOD for varying pose initialization error in rotation (left) and translation (right).}
    \label{fig:pose}
\end{figure}

\textbf{Number of Refinement Iterations: } As indicated in Table \ref{tab:iterations}, SporeAgent significantly improves the pose accuracy within the first few iterations. The learned policy quickly resolves large displacements by selection of the highest magnitude step size. This is also illustrated by the ablation over initial translation errors in Figure \ref{fig:pose} (right). Later iterations further improve alignment.

\textbf{Influence of Initial Pose: }To evaluate robustness to the initialization, we apply a random error of varying magnitude on-top of the ground-truth pose. Rotation and translation errors are evaluated separately with the other kept constant at $10deg$ and $0.1$ units (normalized space), respectively. The random errors are generated as for the training augmentation. A unit vector is randomly and uniformly sampled and interpreted as rotation axis or translation direction. The corresponding rotation angle or translation distance are randomly uniform. %
The AD recalls for varying initial errors are shown in Figure \ref{fig:pose}. The results highlight the robustness of SporeAgent to translation errors. We conjecture that the normalized representation, with a centered target point cloud, simplifies the correction of solely translational offsets. Towards an error of $3$ units in this normalized space, the largest step size of $0.27$ (in combination with at most $10$ iterations) becomes a limiting factor. Rotation errors affect performance more heavily as a partial source may be rotated to align with similar regions of the target point cloud. Yet, up to about $30deg$, the recall for a threshold of $0.10d$ is barely affected.

\subsection{Additional Visualizations}\label{sec:visualization}
The following visualizations aim to illustrate key concepts of SporeAgent by qualitative examples.

\begin{figure}
    \centering
    \includegraphics[trim=0 12 0 0, clip,width=0.5\linewidth]{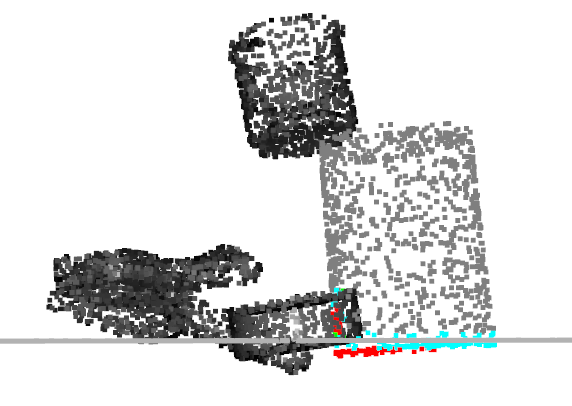}\includegraphics[width=0.5\linewidth]{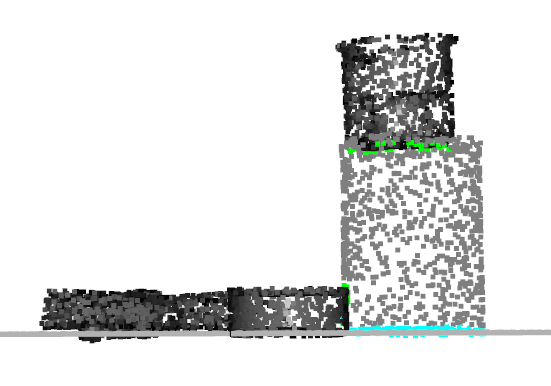}
    \caption{Initial scene representation (left) and refined poses (right). Critical points for one target object (gray) are shown -- intersecting (red), contact (green) and supported (cyan).}
    \label{fig:scene}
\end{figure}

\textbf{Visualizing the Scene Representation: }Figure \ref{fig:scene} shows the target point clouds in the scene representation for a frame in YCBV. Critical points for a queried object (the coffee can, shown in gray) are indicated. Under initial poses (left), the object would intersect with the plane and a neighboring object (shown red). The supported points (cyan) would span a supporting polygon sufficient for static stability. But we define non-intersecting and non-floating as a precondition for plausibility and hence the object pose is considered implausible under its current pose within the scene. The remaining objects in the scene are processed analogously. %
After refinement using SporeAgent (Figure \ref{fig:scene} right), these implausibilities are resolved. Objects are resting on the supporting plane and no longer intersect (subject to slack parameter $\varepsilon$). Contacts (green) of the queried object with the object resting on-top of it are considered non-supported, since the surface normals of the neighboring object near the contacts are pointing in gravity direction.

\begin{figure}
    \centering
    \includegraphics[trim=1100 0 0 0, clip,width=0.5\linewidth]{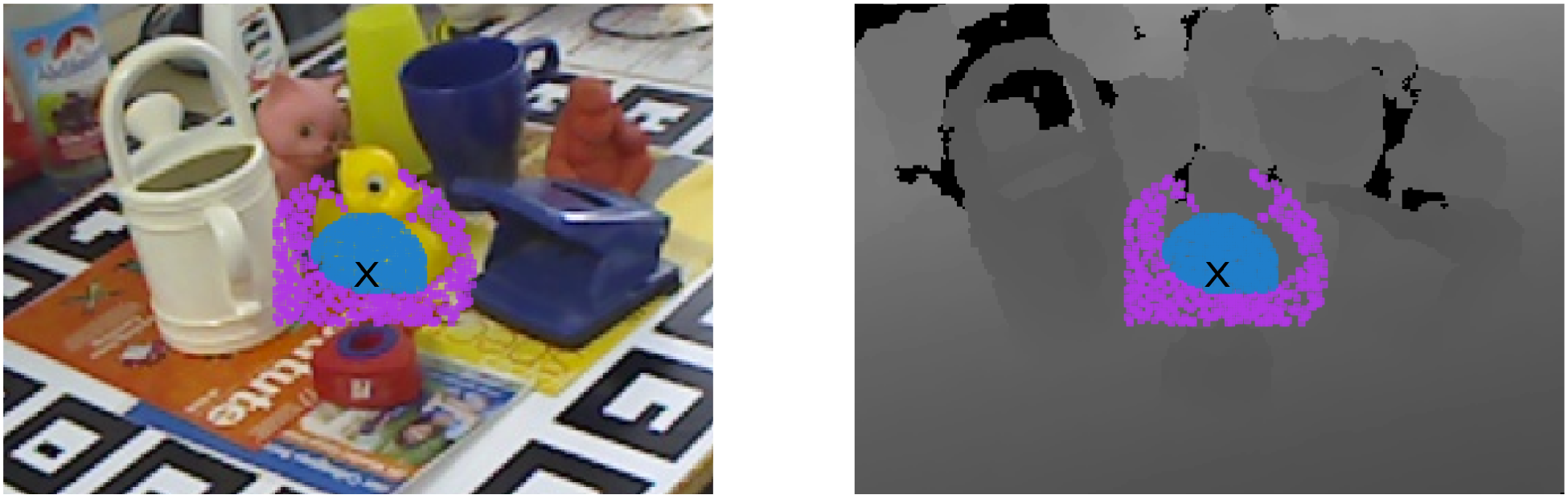}\includegraphics[trim=1100 0 0 0, clip,width=0.5\linewidth]{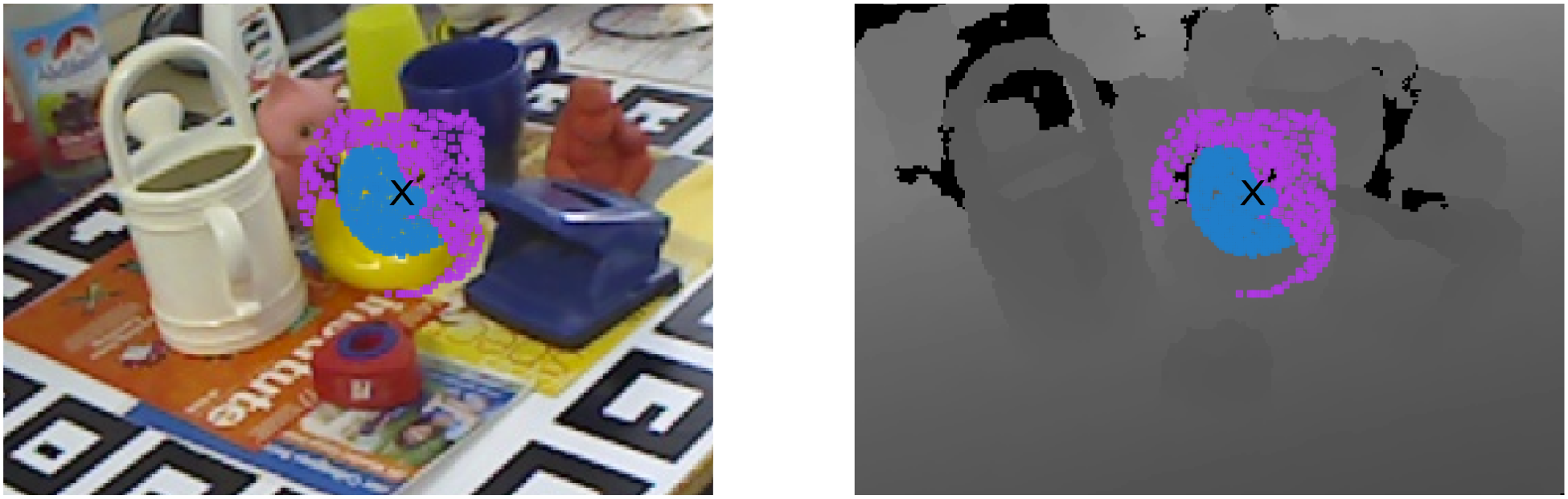}
    \caption{Segmentation augmentation during training with $p=50\%$ foreground samples. Selected center (cross), sampled foreground (blue) and sampled background (purple). Background samples are limited to a bounding box around the target object.}
    \label{fig:segmentation}
\end{figure}

\textbf{Visualizing the Segmentation Augmentation: }The augmentation of the instance segmentation during training is visualized in Figure \ref{fig:segmentation}. The foreground (blue) and background (purple) are determined using the ground-truth visibility mask, with the background restricted to the bounding box around the mask. Both regions are pre-sampled to an equal number of points. Per sample, the augmentation randomly selects one of the foreground pixels (cross) and determines its nearest neighbors in image space. Depending on a uniformly-random fraction $p$ and a total number of points to sample $n$, the $\lceil pn \rceil$ nearest neighbors in the foreground and $\lfloor (1-p)n \rfloor$ nearest neighbors in the background are sampled. This results in a coherent foreground patch, simulating occlusion or a too small mask for $p<1$. The background patch will consist of a part that is coherent with the foreground patch (too large mask, bleeding-out into the surrounding of the object) and a part that fades farther into the surrounding (outliers). As shown in the experiments on YCBV, where training and test scenes are different, this approach allows to learn which points to ignore rather than learning specific scene surroundings.

\begin{figure*}[t]
    \centering
    \begin{tabular}{>{\centering\arraybackslash}p{0.173\linewidth}>{\centering\arraybackslash}p{0.173\linewidth}>{\centering\arraybackslash}p{0.173\linewidth}>{\centering\arraybackslash}p{0.173\linewidth}>{\centering\arraybackslash}p{0.19\linewidth}}
    Init & Step 1 & Step 2 & Result & Observation
    \end{tabular}\\
    \includegraphics[width=0.19\linewidth]{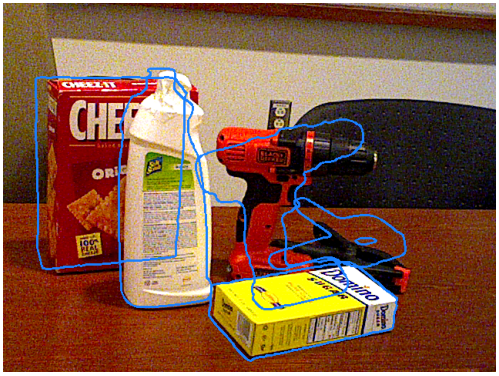} 
    \includegraphics[width=0.19\linewidth]{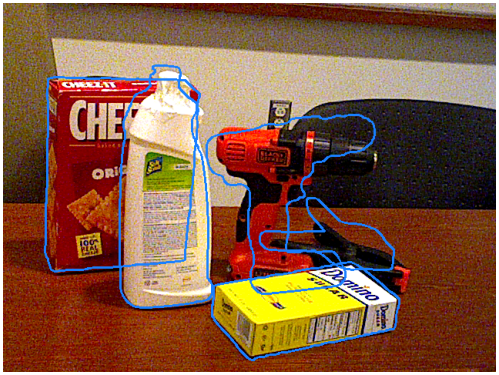} 
    \includegraphics[width=0.19\linewidth]{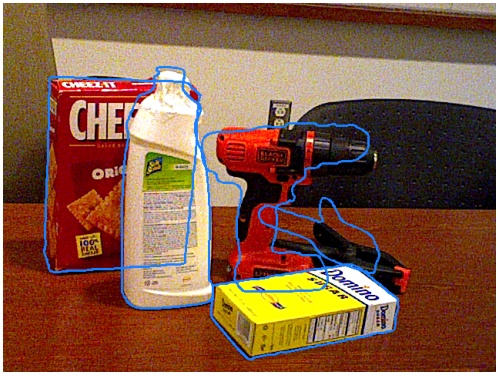} 
    \includegraphics[width=0.19\linewidth]{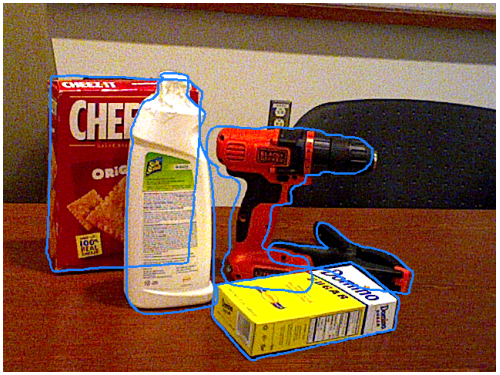} \hspace{1ex} 
    \includegraphics[width=0.19\linewidth]{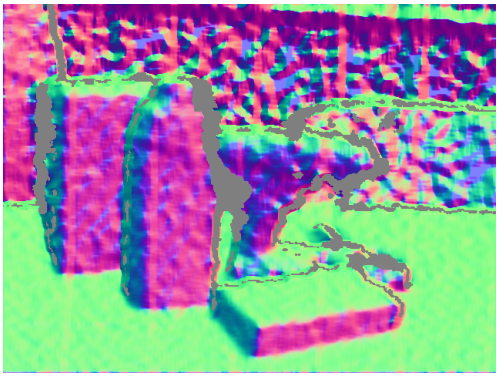}\\
    \includegraphics[width=0.19\linewidth]{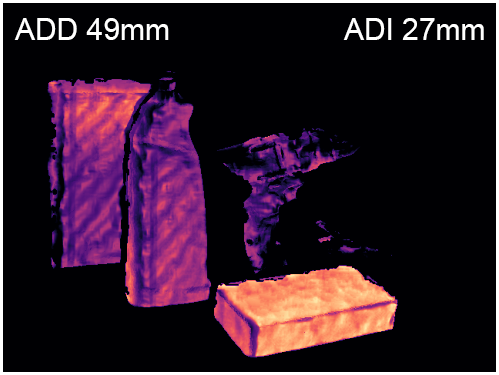} 
    \includegraphics[width=0.19\linewidth]{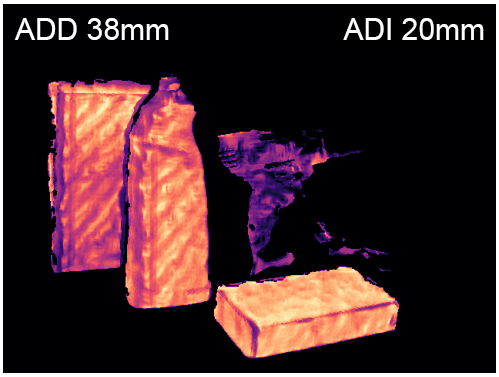} 
    \includegraphics[width=0.19\linewidth]{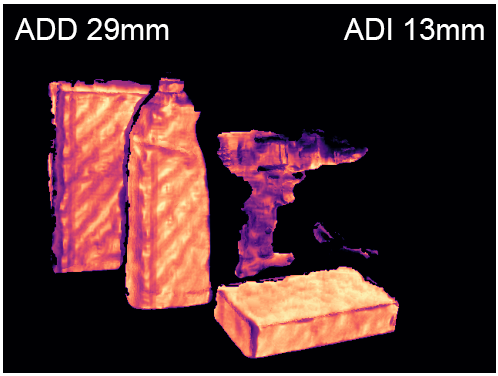} 
    \includegraphics[width=0.19\linewidth]{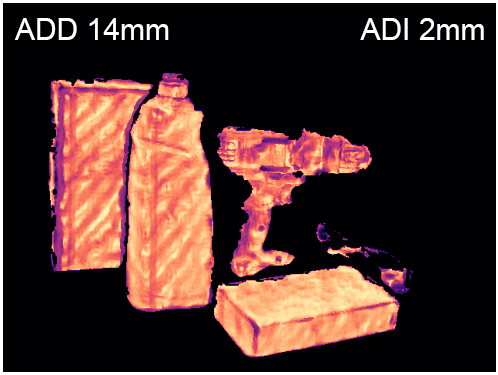} \hspace{1ex}
    \includegraphics[width=0.19\linewidth]{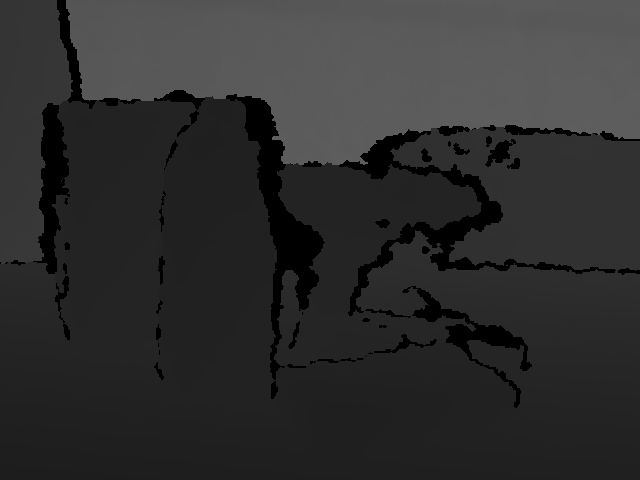}
    \caption{Qualitative step-by-step example with (from left to right) initialization, first two steps and the final result after 10 iterations. The inputs -- the observed normal and depth image -- are shown in the last column. In the corresponding error images used for scoring (bottom row) a brighter color indicates a higher alignment between estimate and observation. ADD/ADI are the mean over per-object distances.}
    \label{fig:stepwise}
\end{figure*}

\textbf{Visualizing the Pose Scoring: }The scoring of pose estimates with respect to the observed frame is illustrated in Figure \ref{fig:stepwise}. The estimates are visualized by outlines (blue) and the corresponding per-pixel score as a heatmap. For the observation, the depth and normal image are shown (right). %
As seen in the left-most column in Figure \ref{fig:stepwise}, only the initial pose estimate for the sugar box (yellow, front) results in close alignment, indicated by a brighter color in the heatmap. Already within the first steps of refinement using SporeAgent, alignment is significantly increased. The best poses after the full refinement result in an even finer alignment, also resolving the large initial pose errors for the driller and the clamp (top right in the frame).

\begin{figure}
    \centering
    \includegraphics[width=0.5\linewidth]{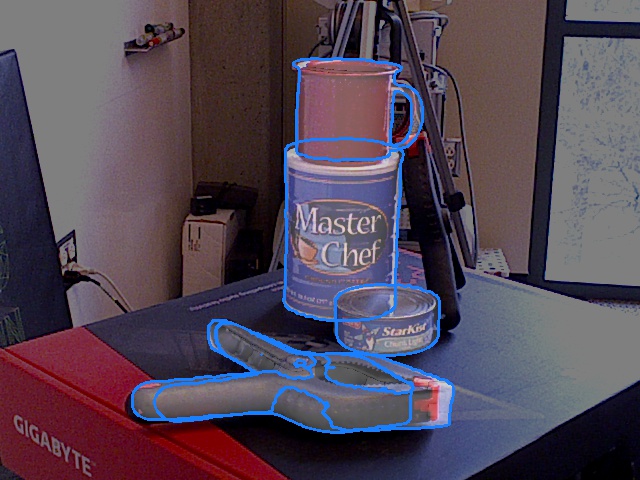}\includegraphics[width=0.5\linewidth]{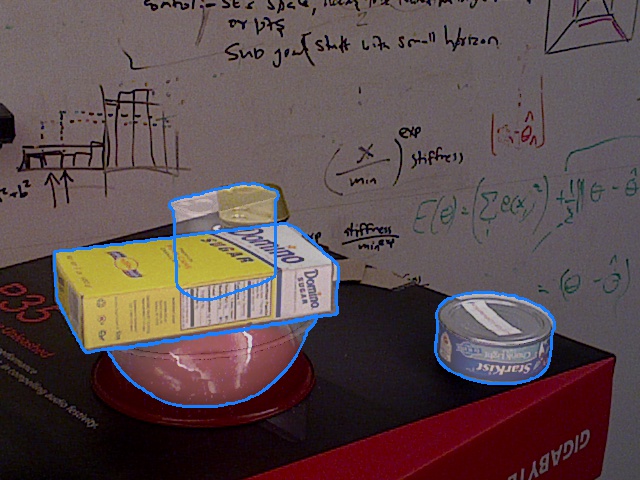}
    \caption{Failure cases on YCB-VIDEO. Best viewed digitally.}
    \label{fig:errors}
\end{figure}

\subsection{Analysis of Failure Cases and Future Work}\label{sec:errors}
Nevertheless, we observe specific systematic failure cases that could not be resolved by considering physical plausibility. For example, as shown in Figure \ref{fig:errors}, the two differently sized clamps in YCBV are typically confused by PoseCNN (our initialization) and may thus end-up stuck within one another. Similarly, the bowl in YCBV may be estimated in an upside-down pose due to occlusion. In this scenario, refinement gets stuck between the plane and the scene object resting on the bowl. We hypothesize that such systematic errors can be best addressed by tighter integration of detection, pose estimation and refinement, as for example proposed by Labb{\'e} et al. \cite{labbe2020cosypose} for CosyPose.

In addition, when inaccurate segmentation, occlusion or missing depth values (e.g., of metallic surfaces like caps of cans) decimate the number of foreground points in the source point cloud too much, accuracy of our approach significantly drops as the remainder of the source mostly consists of background points. Additional consideration of color information might lessen this issue.

Further robustness may be achieved by using multiple pose hypotheses per instance, as in related work \cite{labbe2020cosypose,mitash2018mcts,bauer2020verefine}. For example, in the case of the confused clamp classes, one hypothesis per class might be refined and the most plausible one selected using the rendering-based scoring.

\section{Conclusion}\label{sec:conclusion}
We present \textit{SporeAgent}, a reinforcement learning approach to object pose refinement. SporeAgent jointly refines the poses of multiple objects in an observed depth frame in parallel. By considering object symmetries and physical plausibility of the scene, we achieve state-of-the-art results as shown in our evaluation on the LINEMOD and YCB-VIDEO datasets. The provided ablation study illustrates the benefit of each proposed part and our analysis of failure cases motivates future improvements towards further robustness to noise and inaccuracy in detection, segmentation and initialization.

\textbf{Acknowledgements:} This work was supported by the TU Wien Doctoral College TrustRobots and the Austrian Science Fund (FWF) under grant agreements \mbox{No. I3968-N30} HEAP and No. I3969-N30 InDex.

{\small
\bibliographystyle{ieee_fullname}
\bibliography{references}

\begin{thebibliography}{10}\itemsep=-1pt

\bibitem{aoki2019pointnetlk}
Yasuhiro Aoki, Hunter Goforth, Rangaprasad~Arun Srivatsan, and Simon Lucey.
\newblock Pointnetlk: Robust \& efficient point cloud registration using
  pointnet.
\newblock In {\em IEEE Conf. Comput. Vis. Pattern Recog.}, pages 7163--7172,
  2019.

\bibitem{bauer2020eccvw}
Dominik Bauer, Timothy Patten, and Markus Vincze.
\newblock Physical plausibility of 6d pose estimates in scenes of static rigid
  objects.
\newblock In {\em Eur. Conf. Comput. Vis. Workshops}, pages 648--662, 2020.

\bibitem{bauer2020verefine}
Dominik Bauer, Timothy Patten, and Markus Vincze.
\newblock Verefine: Integrating object pose verification with physics-guided
  iterative refinement.
\newblock {\em IEEE Robot. Autom. Lett.}, 5(3):4289--4296, 2020.

\bibitem{bauer2021reagent}
Dominik Bauer, Timothy Patten, and Markus Vincze.
\newblock Reagent: Point cloud registration using imitation and reinforcement
  learning.
\newblock In {\em IEEE Conf. Comput. Vis. Pattern Recog.}, pages 14586--14594,
  2021.

\bibitem{besl1992icp}
Paul~J. Besl and Neil~D. McKay.
\newblock A method for registration of 3-d shapes.
\newblock {\em IEEE Trans. Pattern Anal. Mach. Intell.}, 14(2):239--256, 1992.

\bibitem{brachmann2016uncertainty}
Eric Brachmann, Frank Michel, Alexander Krull, Michael~Ying Yang, Stefan
  Gumhold, et~al.
\newblock Uncertainty-driven 6d pose estimation of objects and scenes from a
  single rgb image.
\newblock In {\em IEEE Conf. Comput. Vis. Pattern Recog.}, pages 3364--3372,
  2016.

\bibitem{busam2020moveit}
Benjamin Busam, Hyun~Jun Jung, and Nassir Navab.
\newblock I like to move it: 6d pose estimation as an action decision process.
\newblock {\em arXiv preprint arXiv:2009.12678}, 2020.

\bibitem{calli2015ycb}
Berk Calli, Arjun Singh, Aaron Walsman, Siddhartha Srinivasa, Pieter Abbeel,
  and Aaron~M Dollar.
\newblock The ycb object and model set: Towards common benchmarks for
  manipulation research.
\newblock In {\em Int. Conf. on Adv. Robot.}, pages 510--517. IEEE, 2015.

\bibitem{chen1992p2pl}
Yang Chen and G{\'e}rard Medioni.
\newblock Object modelling by registration of multiple range images.
\newblock {\em Image and Vis. Comput.}, 10(3):145--155, 1992.

\bibitem{choy2020dgr}
Christopher Choy, Wei Dong, and Vladlen Koltun.
\newblock Deep global registration.
\newblock In {\em IEEE Conf. Comput. Vis. Pattern Recog.}, pages 2514--2523,
  2020.

\bibitem{deng2021poserbpf}
Xinke Deng, Arsalan Mousavian, Yu Xiang, Fei Xia, Timothy Bretl, and Dieter
  Fox.
\newblock Poserbpf: A rao--blackwellized particle filter for 6-d object pose
  tracking.
\newblock {\em IEEE Trans. Robot.}, 2021.

\bibitem{hinterstoisser2012adi}
Stefan Hinterstoisser, Vincent Lepetit, Slobodan Ilic, Stefan Holzer, Gary
  Bradski, Kurt Konolige, and Nassir Navab.
\newblock Model based training, detection and pose estimation of texture-less
  3d objects in heavily cluttered scenes.
\newblock In {\em ACCV}, pages 548--562, 2012.

\bibitem{hodan2020bop}
Tom{\'a}{\v{s}} Hoda{\v{n}}, Martin Sundermeyer, Bertram Drost, Yann Labb{\'e},
  Eric Brachmann, Frank Michel, Carsten Rother, and Ji{\v{r}}{\'i} Matas.
\newblock {BOP} challenge 2020 on {6D} object localization.
\newblock {\em Eur. Conf. Comput. Vis. Workshops}, 2020.

\bibitem{krull2017poseagent}
Alexander Krull, Eric Brachmann, Sebastian Nowozin, Frank Michel, Jamie
  Shotton, and Carsten Rother.
\newblock Poseagent: Budget-constrained 6d object pose estimation via
  reinforcement learning.
\newblock In {\em IEEE Conf. Comput. Vis. Pattern Recog.}, pages 6702--6710,
  2017.

\bibitem{labbe2020cosypose}
Yann Labb{\'e}, Justin Carpentier, Mathieu Aubry, and Josef Sivic.
\newblock Cosypose: Consistent multi-view multi-object 6d pose estimation.
\newblock In {\em Eur. Conf. Comput. Vis.}, pages 574--591, 2020.

\bibitem{li2018deepim}
Yi Li, Gu Wang, Xiangyang Ji, Yu Xiang, and Dieter Fox.
\newblock Deepim: Deep iterative matching for 6d pose estimation.
\newblock In {\em Eur. Conf. Comput. Vis.}, pages 683--698, 2018.

\bibitem{mitash2018mcts}
Chaitanya Mitash, Abdeslam Boularias, and Kostas~E Bekris.
\newblock Improving 6d pose estimation of objects in clutter via physics-aware
  monte carlo tree search.
\newblock In {\em Int. Conf. Robot. Autom.}, pages 3331--3338, 2018.

\bibitem{qi2017pointnet}
Charles~R Qi, Hao Su, Kaichun Mo, and Leonidas~J Guibas.
\newblock Pointnet: Deep learning on point sets for 3d classification and
  segmentation.
\newblock In {\em IEEE Conf. Comput. Vis. Pattern Recog.}, pages 652--660,
  2017.

\bibitem{rad2017bb8}
Mahdi Rad and Vincent Lepetit.
\newblock Bb8: A scalable, accurate, robust to partial occlusion method for
  predicting the 3d poses of challenging objects without using depth.
\newblock In {\em Int. Conf. Comput. Vis.}, pages 3828--3836, 2017.

\bibitem{schulman2017ppo}
John Schulman, Filip Wolski, Prafulla Dhariwal, Alec Radford, and Oleg Klimov.
\newblock Proximal policy optimization algorithms.
\newblock {\em arXiv preprint arXiv:1707.06347}, 2017.

\bibitem{shao2020pfrl}
Jianzhun Shao, Yuhang Jiang, Gu Wang, Zhigang Li, and Xiangyang Ji.
\newblock Pfrl: Pose-free reinforcement learning for 6d pose estimation.
\newblock In {\em IEEE Conf. Comput. Vis. Pattern Recog.}, pages 11454--11463,
  2020.

\bibitem{tekin2018real}
Bugra Tekin, Sudipta~N Sinha, and Pascal Fua.
\newblock Real-time seamless single shot 6d object pose prediction.
\newblock In {\em IEEE Conf. Comput. Vis. Pattern Recog.}, pages 292--301,
  2018.

\bibitem{wada2020morefusion}
Kentaro Wada, Edgar Sucar, Stephen James, Daniel Lenton, and Andrew~J Davison.
\newblock Morefusion: multi-object reasoning for 6d pose estimation from
  volumetric fusion.
\newblock In {\em IEEE Conf. Comput. Vis. Pattern Recog.}, pages 14540--14549,
  2020.

\bibitem{wang2019densefusion}
Chen Wang, Danfei Xu, Yuke Zhu, Roberto Mart{\'\i}n-Mart{\'\i}n, Cewu Lu, Li
  Fei-Fei, and Silvio Savarese.
\newblock Densefusion: 6d object pose estimation by iterative dense fusion.
\newblock In {\em IEEE Conf. Comput. Vis. Pattern Recog.}, pages 3343--3352,
  2019.

\bibitem{wang2019dcp}
Yue Wang and Justin~M Solomon.
\newblock Deep closest point: Learning representations for point cloud
  registration.
\newblock In {\em Int. Conf. Comput. Vis.}, pages 3523--3532, 2019.

\bibitem{xiang2017posecnn}
Yu Xiang, Tanner Schmidt, Venkatraman Narayanan, and Dieter Fox.
\newblock Posecnn: A convolutional neural network for 6d object pose estimation
  in cluttered scenes.
\newblock In {\em Robot.: Sci. Syst.}, 2018.

\bibitem{yew2020rpmnet}
Zi~Jian Yew and Gim~Hee Lee.
\newblock Rpm-net: Robust point matching using learned features.
\newblock In {\em IEEE Conf. Comput. Vis. Pattern Recog.}, pages 11824--11833,
  2020.

\bibitem{yuan2020deepgmr}
Wentao Yuan, Ben Eckart, Kihwan Kim, Varun Jampani, Dieter Fox, and Jan Kautz.
\newblock Deepgmr: Learning latent gaussian mixture models for registration.
\newblock {\em arXiv preprint arXiv:2008.09088}, 2020.

\bibitem{zakharov2019dpod}
Sergey Zakharov, Ivan Shugurov, and Slobodan Ilic.
\newblock Dpod: 6d pose object detector and refiner.
\newblock In {\em Int. Conf. Comput. Vis.}, pages 1941--1950, 2019.

\bibitem{open3d}
Qian-Yi Zhou, Jaesik Park, and Vladlen Koltun.
\newblock Open3d: A modern library for 3d data processing.
\newblock {\em arXiv preprint arXiv:1801.09847}, 2018.

\end{thebibliography}
}

\end{document}